\documentclass[10pt,twocolumn,letterpaper]{article}

\usepackage{iccv}
\usepackage{times}
\usepackage{epsfig}
\usepackage{graphicx}
\usepackage{amsmath}
\usepackage{amssymb}
\usepackage{mathtools}
\usepackage[table,x11names,dvipsnames]{xcolor}
\usepackage{booktabs}
\usepackage{adjustbox}
\usepackage{caption}
\usepackage{listings}
\usepackage{algorithm2e}
\usepackage{pythonhighlight} 

\lstnewenvironment{pythonic}[1][]{\lstset{style=mypython, frame=none, #1}}{}

\usepackage[pagebackref=true,breaklinks=true,letterpaper=true,colorlinks,bookmarks=false]{hyperref}
\usepackage[capitalize]{cleveref}

 \iccvfinalcopy 

\def\iccvPaperID{26} 

\ificcvfinal\fi

\begin{document}

\title{An Empirical Analysis of Range for 3D Object Detection}  
\author{Neehar Peri$^1$, Mengtian Li$^1$, Benjamin Wilson$^2$, Yu-Xiong Wang$^3$, James Hays$^2$, Deva Ramanan$^1$\\
Carnegie Mellon University$^1$, Georgia Institute of Technology$^2$, University of Illinois Urbana-Champaign$^3$ \\
{\tt\small \{nperi,mtli,deva\}@andrew.cmu.edu, \{benjaminrwilson,hays\}@gatech.edu, yxw@illinois.edu}
}

\ificcvfinal\thispagestyle{empty}\fi

\twocolumn[{%
\maketitle
\vspace{-2.5em}
\centering
\includegraphics[width=0.85\linewidth, clip, trim={0cm 0cm 10cm 0cm}]{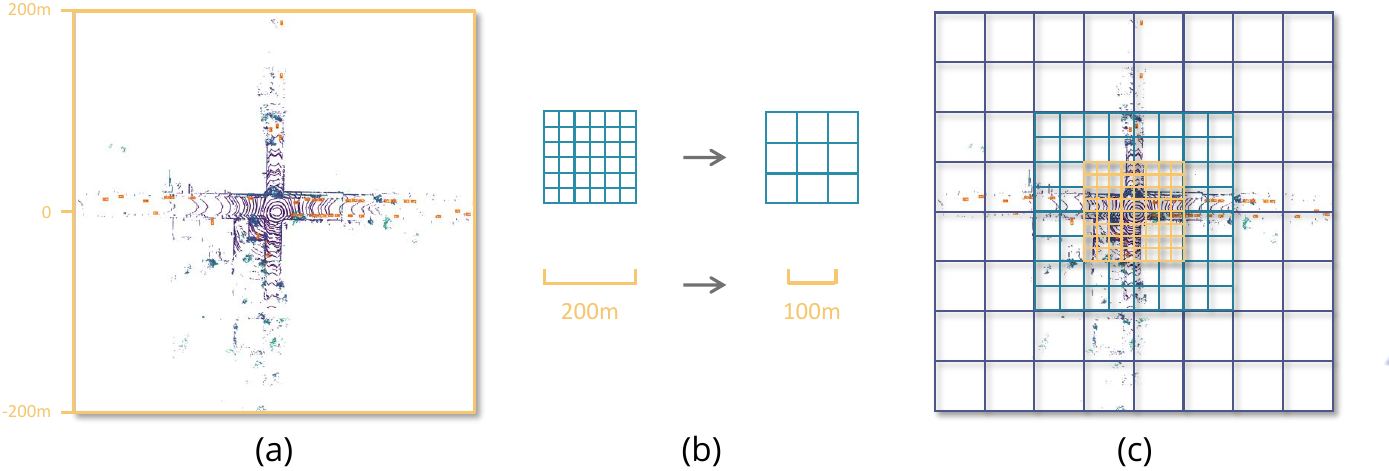}
\vspace{0.25em}
\captionof{figure}{
(a) Existing 3D LiDAR detectors 
struggle to detect far away objects (\eg, $300$m) due to time and compute constraints. (b) We explore two ways to reduce compute: adopt a coarser grid and limit the processing range. Somewhat surprisingly, we find that range is the single most effective ``knob" for trading off accuracy-vs-latency, implying that detectors {\em should} ``give up" on far-field detection to manage compute budgets.  Moreoever, we find that near-range detectors can exploit finer voxel sizes for higher resolution processing, while far-range detectors benefit from larger voxels. We denote models optimized for specific ranges as \textit{range experts}. (c) To avoid blindly giving on far-field objects, we simply combine range experts by ensembling; e.g., combine 0-50m detections from the 50m expert with 50-100m detections from the 100m expert. Despite improved performance, the runtime of this naive range ensemble increases linearly with the number of range experts. To address this, we introduce near-far range-ensembles, which take inspiration from hierarchical controllers to run near-field detectors (for near-term collision avoidance) at a higher rate than far-field detectors (for long-horizon planning).  
}
\vspace{0.5em}
\label{fig:teaser}
}]



\begin{abstract}
LiDAR-based 3D detection plays a vital role in autonomous navigation. Surprisingly, although autonomous vehicles (AVs) must detect both near-field objects (for collision avoidance) and far-field objects (for longer-term planning), contemporary benchmarks focus only on near-field 3D detection. However, AVs must detect far-field objects for safe navigation. In this paper, we present an empirical analysis of far-field 3D detection using the long-range detection dataset Argoverse 2.0 to better understand the problem, and share the following insight: near-field LiDAR measurements are dense and optimally encoded by small voxels, while far-field measurements are sparse and are better encoded with large voxels. We exploit this observation to build a collection of range experts tuned for near-vs-far field detection, and propose simple techniques to efficiently ensemble models for long-range detection that improve efficiency by 33\% and boost accuracy by 3.2\% CDS.

\end{abstract}

\section{Introduction}
3D object detection is a critical component of the autonomy stack.
Despite the maturity of methods in existing literature,
most treat detection range as a constant instead of an adjustable hyperparameter \cite{yan2018second,lang2019pointpillars,yin2021center}, likely because existing benchmarks primarily evaluate near-field detections. Motivated by highway driving and long-horizon planning, we present an empirical analysis of far-range perception and share insights that are widely applicable across model architectures. Contemporary solutions for near-field 3D detection make use of 3D voxel representations, often encoded with a bird’s-eye view (BEV) feature map. While quite intuitive, such representations scale quadratically with the spatial range of the map. We find that a primary challenge for effectively addressing long-range 3D detection is managing compute and latency demands.
\begin{figure*}[t]
\centering
\includegraphics[width=0.80\linewidth, clip, trim={2cm 4cm 4cm 6cm}]{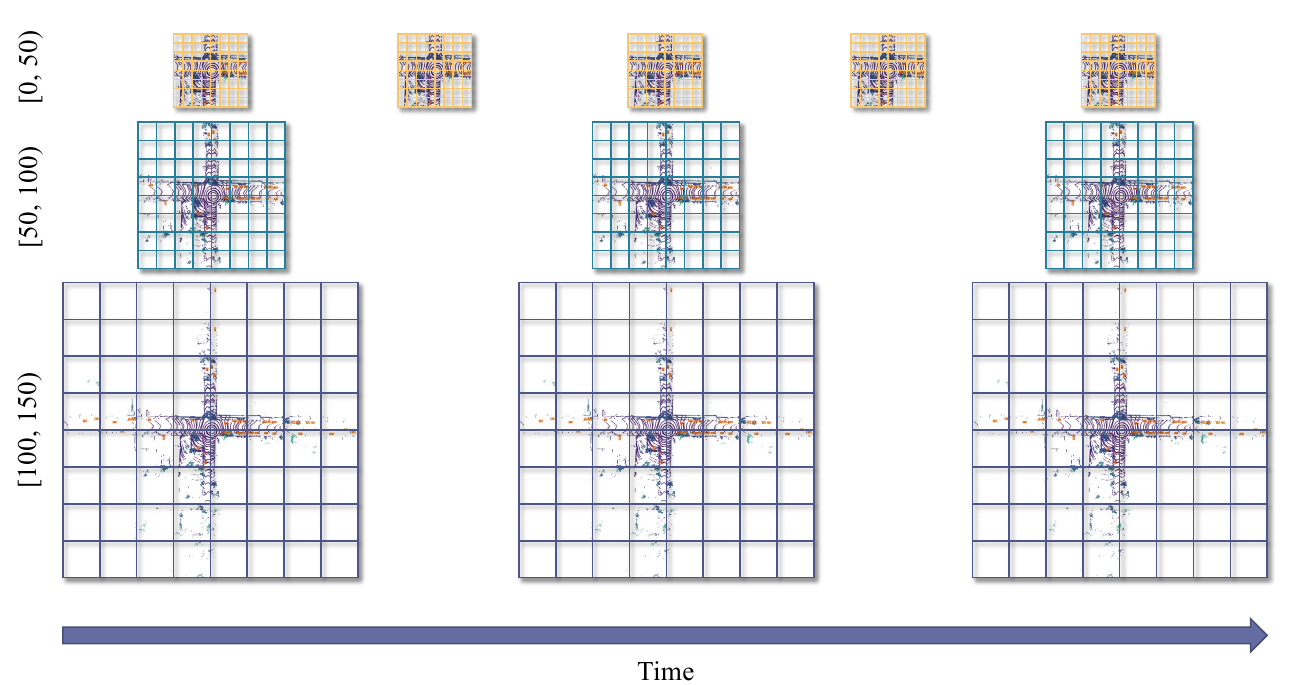}
\caption{{\bf Near-Far Ensemble}. We can achieve considerable efficiency gains by processing different range experts \textit{asynchronously} at different frequencies. Importantly, near-field detectors need to be processed more frequently than far field detectors for autonomous navigation. We run the high resolution near-field model at every timestamp and process the medium and long-range models at lower frequencies. To estimate object locations for frames without far-field processing, we forecast previous detections using a constant velocity model. By simply running far-field detectors less often, we achieve a 33\% increase in efficiency over the naive range ensemble. 
}
\label{fig:near_far}
\end{figure*}
{\bf Accuracy-vs-Latency.} In this paper, we study different factors for trading off compute-vs-latency, like voxel resolution and detection range, 
in the context of bird's-eye view (BEV) based 3D detection. BEV-based detectors operate on a dense 2D BEV feature map 
whose spatial dimensions are directly determined by the processing range and the voxel size (grid density). 
Interestingly, the existing literature on {\em 2D} detection has converged on image resolution and backbone depth as the handy ``knobs'' for trading off accuracy-vs-latency~\cite{tan2020efficientdet,Li2020StreamingP,wang2021scaled}. 
We revisit these questions in the context of 3D detection, and find somewhat surprisingly that {\em range} is an even more effective parameter for trading off these quantities (c.f. Fig. \ref{fig:teaser}).
For example, we show that even if the sensor (dataset) includes object annotations up to 150m, optimal accuracy-vs-latency tradeoffs may be achieved by artificially limiting the range of the model to 100m, essentially ``giving up" on far-field detections during training. We posit that this is due to the distribution of annotations (e.g. fewer objects are labeled in the far-field).

{\bf Range Ensemble.} One interesting by-product of giving-up on the far-field is that the additional compute can be re-allocated to the near-field via smaller (higher resolution) voxels. Our analysis reveals that bird's-eye view (BEV) representations can be tuned for particular ranges by adjusting other hyperparameters such as voxel resolution. We denote models optimized for specific ranges as \textit{range experts}.
Ultimately, we would like to avoid giving up on the far-field as it is important for highway driving and long-horizon planning. To avoid doing so, we simply combine range experts by ensembling; e.g., combine 0-50m detections from the 50m expert with 50-100m detections from the 100m expert. We find such an ensemble greatly boosts detection accuracy. Perhaps unsurprisingly, such an architecture is performant because it exploits a well-known but under-emphasized property of LiDAR: {\em farther range implies greater sparsity}. 

{\bf LiDAR Sparsity.} Interestingly, prior work \cite{zhang2020polarnet, rapoport2021s, jiang2022polarformer} has exploited sparsity in the context of spherical voxelization or range-view processing. While performant for tasks such as semantic segmentation
~\cite{zhu2021cylindrical}, most SOTA architectures for 3D cuboid detection still make use of rectilinear voxel grids. One reason may be that spherical warping introduces perspective distortions that warp far-field objects, making it difficult to explicitly tune range. In contrast, our range ensemble can be seen as a rectilinear approximation of spherical voxelization that {\em avoids} voxel distortion. Moreover, due to the popularity of rectilinear detectors, there exist more mature methods for data augmentation~\cite{fan2021rangedet} and temporal fusion, either at the sensor level \cite{yan2018second,lang2019pointpillars,yin2021center, peri2022futuredet} or at the feature level~\cite{luo2018fast,Huang2020AnLA}, which is essential for long-range detection. Due to the strong empirical performance of 3D BEV detectors~\cite{yin2021center,zhu2019class,Liu2022BEVFusionMM,bai2022transfusion}, we argue that improving their range-efficiency will be increasingly important as LiDAR sensors themselves increase in range and density. 


{\bf Near-Far Ensembles.} Finally, we demonstrate that one can trivially speed up a multi-range ensemble via range-specific {\em asynchronous} processing. We take inspiration from hierarchical ``slow-fast" planners that run a low-frequency planner together with a high-frequency reactive controller. From a perception perspective, autonomous agents need to quickly react to near-field objects (that represent potential collisions), while far-field objects may be used for more strategic long-term planning.  
Concretely, we run near-range experts at high frequency and run far-range experts at a lower frequency (Fig.~\ref{fig:near_far}) Our results highlight an interesting observation: sometimes it is more effective to {\em forecast} the location of a far-field object from a previous frame's detection than to directly process the far-field of the current frame. One reason is that the object may have appeared in the near-field of the previous frame, making it far easier to detect. We find that near-far ensembles reduce runtime by 33\% with little performance decrease.


{

We summarize our contributions as follows:
\begin{enumerate}
    \item We study the impact of range as a tunable parameter for 3D object detection. We draw analogy to image resolution and find the surprising conclusion that the best solution to optimize the accuracy-vs-latency tradeoff is to "give up" on far-field detection.
    \item We study how detectors can generalize across ranges (e.g. train on 50m, but deploy at 100m) due to fully convolutional processing. We find that certain architectural design choices, such as voxel encoding and detector head design greatly impact across-range generalization.
    \item We present a simple extension of range ensembles that takes inspiration from hierarchical controllers by running near-field detectors at higher frequency and far-field detectors at lower frequency. We show improved efficiency over the naive range-ensemble, reducing latency by 33\%!
\end{enumerate}

\section{Related Work}



3D detection models can be roughly categorized as: bird's-eye view, voxel-grid, point, graph, and range-view representation models. Unlike 2D images, point clouds are amenable to a number of different representations, each with distinct advantages and disadvantages, particularly in the context of long-range detection.

\textbf{Bird's-eye view Representations.}
3D perception using \emph{2D} convolutions enables fast, efficient feature aggregation due to mature, highly optimized kernels available in open-source libraries. However, these methods must be carefully designed to encode geometric information in the height dimension.
PointPillars \cite{lang2019pointpillars} represents a point cloud as a ``pseudo-image'', applying a PointNet \cite{qi2017pointnet} encoding to a set of sparse pillars in the BEV. 
MV3D \cite{chen2017multi} explores a multi-sensor fusion model which consists of a bird's-eye and range view of LiDAR sensor data, and RGB imagery. However, ego-centric point clouds are \emph{not} dense in the BEV, which consequently wastes both memory and computation. Specialized sparse operators \emph{may} address the issues of density, but are often not as well-tuned as 2D convolutions for GPU-based computation. Further, we find that different implementations of sparse convolutions can have a significant impact on latency.

\textbf{Voxel-grid Representations.}
3D convolution provides rich, expressive geometric features at the cost of a cubic run-time w.r.t. the quantized grid dimensions leading to considerable compute challenges. 
VoxelNet \cite{zhou2018voxelnet} introduced the first end-to-end learning approach for 3D object detection by augmenting point features with positional encodings within a voxel-grid. SECOND \cite{yan2018second} exploits the sparsity of a point cloud through 3D \emph{sparse} convolutions, greatly improving run-time to speeds suitable for real-time applications. \cite{tang2020searching} combines voxel and point level processing to exploit the \emph{efficiency} of a regular grid and the \emph{geometric richness} of point-level features. Similar to our work, \cite{zhang2020polarnet,zhu2021cylindrical} emphasize that point clouds are \emph{sparse} at range, leading to an \emph{imbalanced} spatial distribution of points. Prior works address this observation by representing point clouds with polar and cylindrical representations, respectively. However, this can lead to spatial distortions that break the translation equivariance assumed by convolutional filters. Prior work also explores multi-resolution voxel-based approaches. \cite{hu2022point} proposes a density-aware RoI grid pooling module using kernel density estimation and self-attention with point density positional encoding to efficiently voxelize and encode LiDAR points. \cite{kuang2020voxel} proposes a bottom-up multi-scale voxel encoder and a top-down multi-scale feature map aggregator.

\textbf{Graph Representations.}
Graph representations of point clouds encode \emph{dynamic} neighborhoods between points while also permitting a sparse representation in the form of a sparse matrix or adjacency list. PointNet++ \cite{qi2017pointnet++} encodes a point cloud as a hierarchical set of point-feature abstractions for 3D object classification. \cite{wang2019dynamic} introduces the EdgeConv operator which directly aggregates \emph{edge} features of point clouds while maintaining permutation invariance.
Point-RCNN \cite{shi2019pointrcnn} operates directly on point clouds without voxelization, using point-wise feature vectors for bottom up proposal generation. 
\cite{hu2020randla} proposes using random sampling to process large point clouds.
Graph representations oftentimes require \emph{costly} neighborhood computation using k-nearest or fixed-radius nearest neighbors.
Despite their flexible representation, 3D detection models on state-of-the-art leaderboards are still dominated by voxel-grid and BEV based methods \cite{caesar2020nuscenes,sun2020scalability}. 

\textbf{Range-view Representations.}
Range-view refers to the projection of an unordered set of three-dimensional coordinates onto a two-dimensional grid which represents the distance from a \emph{visible} point to the sensor. Unlike voxel-grid or graph representations, range-view is not information-preserving for 3D data, i.e., each sensor return must have a clear line-of-sight between itself and the vantage point.
LaserNet \cite{meyer2019lasernet} combines a range-view representation with probabilistic cuboid encoding for 3D detection. \cite{chai2021point} explores applying different kernels to the range-view image to counteract perspective distortions and large depth gradients w.r.t. to inclination and azimuth. \cite{sun2021rsn} construct a two-stage approach, first performing foreground segmentation in range-view and applying sparse convolutions on the remaining points. However, much like speherical voxelization, range view suffers from perspective distortions and far-away objects have a smaller footprint in a range image.


\section{Approach}
In this section, we propose several approaches for effectively tuning range-experts (Sec. \ref{ssec:range-expert}), efficiently constructing range-ensembles (Sec \ref{ssec:range-ensemble}), and further optimizing ensemble runtime with near-far networks (Sec \ref{ssec:near-far}).

\subsection{Range Experts}
\label{ssec:range-expert}
Detection range is largely considered as a constant in the literature. However, properly tuning the detection range together with voxel size can yield a better performance-latency trade off.
We derive model-families by tuning the range parameter and evaluate the accuracy and latency across different range intervals. Since the baseline detector is fully-convolutional (as are many other detectors), we can run inference at a different range from training. Taking this into consideration, we introduce a new notation of $r_1/s$ → $r_2$, where $r_1$ represents the range the model is trained at, $s$ represents the reciprocal of the voxel size and $r_2$ represents the inference range. Note that the voxel size must remain the same during training and testing.

\textbf{Train-Time Range Masking.} In order to maximize the performance of a far-field range expert, one may naturally assume that allocating model capacity to near-field regions of the LiDAR sweep may negatively impact model performance. Concretely, when training a 100m range expert to detect objects between 50-100m, it seems wasteful to spend processing time on the 0-50m region of the point cloud. In particular, the point density for near-field regions are significantly higher than far-field regions. Additionally, more objects are annotated in the near-field, so we expect that this distributions shift will negatively affect generalization. In fact, the standard practice {\em already} crops out too-far regions for each model (by simply limiting their max range). To address this concern, we train range experts with a masked out ``donut hole'' (c.f. Fig. \ref{fig:donut-hole}) to remove LiDAR points and ground truth annotations outside of the region of interest. Somewhat surprisingly, we find that this ``donut hole'' range masking during training time hurts performance. In practice, it seems that learning to detect near-field cars helps to detect far-field cars.

\textbf{Test-Time Range Masking.}
Although we find that ``donut hole'' range masking hurts performance during training, we find that only applying it during test-time does not affect detection accuracy (c.f. Fig \ref{fig:donut-hole}). In fact, it provides a modest speedup because it makes better use of sparse computation. Note that this speedup is largely dependent on the sparse voxelization encoder. We find that this provides considerable improvement for models using spconv 1.0, but limited improvement for models using spconv 2.0. 

\textbf{Up-sampling Far-Field Objects.} One reason that near-field detections may help improve far-field detections is because there are simply too few objects annotated in the far field. Taking inspiration from long-tailed detection literature \cite{zhu2019class, peri2022lt3d}, we can upsample LiDAR sweeps with more far-field ground truth objects and paste more examples of far-field objects into each sweep. However, we find that this does even worse than range masking at training time. We posit that the distribution of objects seen during training time is considerably different than that see during inference, leading to a significant performance drop. 


\begin{figure}[t]
    \centering
    \includegraphics[width=\linewidth, clip, trim={4cm 10cm 10.5cm 3.5cm}]{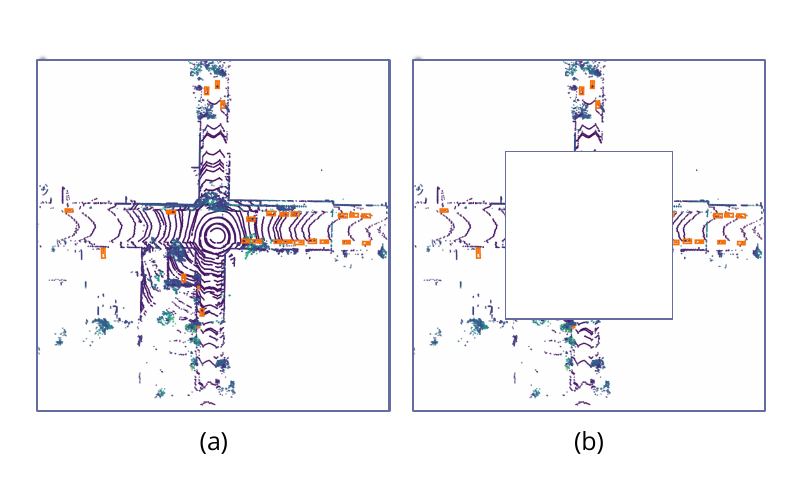}
    \caption{{\bf Range Masking}. We evaluate the effectiveness of range masking when training and evaluating range experts. We find that (a) range masking negatively impacts performance during training. Surprisingly, learning to detect near-range cars improves the detection of far-field cars. However, (b) range masking reduces latency during inference as the additional sparsity increases the efficiency of sparse voxel encoders.}
    \label{fig:donut-hole}
\end{figure}

\subsection{Range Ensembles}
\label{ssec:range-ensemble}
After training each range-expert, we can ensemble their detections by combining predictions from the respective expert models. One strategy is to pool all detections together and perform non-maximal suppression to remove duplicate detections across range experts. However, we find that this approach introduces more false positives and negatively impacts overall performance. Instead, we run each range expert as normal, but simply post-process detections from each range expert such that they only contribute to detecting objects within their tuned range. However, this wastes a considerable amount of compute. Specifically, we must still run our 100m range expert on the dense 0-50m regions of the point cloud, despite discarding these predictions during post processing. Instead, we opt to exploit sparse convolutions to speed up inference using test-time range masking.

\subsection{Near-Far Ensembles}
\label{ssec:near-far}
We explore the idea of near-far ensembles, as shown in Fig. \ref{fig:near_far}, to considerably improve the run-time of a range ensembles. We can achieve these efficiency gains by processing different range experts \textit{asynchronously} at different frequencies. Intuitively, we care more about changes to our near-field to avoid immediate collisions and can update the far-field region less frequently for longer-term planning. We run the high resolution near-field model at every timestamp and process the long-range models at every other timesetep. To estimate object locations for frames without far-field processing, we forecast previous detections using a constant velocity model. We are able to predict per-object velocity estimates from our models because the input to our detectors is a stack of aggregated LiDAR sweeps which implicitly encodes object motion. This constant velocity forecast, which simply updates the past object location using the predicted object velocity estimate (e.g. \pyth{box.center+=box.velocity*time_delta}), is reasonable because we are only forecasting 0.5 seconds into the future. Intuitively, all object motion can be linearized given a sufficiently small time delta. Importantly, since all detections are in the ego-vehicle coordinate frame, forecasting past detections into the current frame requires ego-motion compensation between frames. We present simplified python-like pseudo-code in Alg. \ref{alg:near-far}.  

\RestyleAlgo{ruled}

\SetKwComment{Comment}{/* }{ */}

\begin{algorithm}[hbt!]
\caption{We present python style pseudo-code for the near-far range ensemble. Concretely, we run the near-range expert model at every time step, but only run the far-range expert every $freq$ timesteps (default is 2). We assume that the forecaster compensates for ego-motion.}\label{alg:near-far} 

\begin{pythonic}
#near_expert: Near-range expert detector
#far_expert: Far-range expert detector
#freq: Frequency of of far-range detector
#donut_crop: Removes near-range lidar points
#forecast: Constant-velocity forecast
#dets: Dict[List] of detections

for time, lidar_sweep in enumerate(data):
    # Run near-field range expert
    near_dets = near_expert(lidar_sweep)
    
    if time % freq == 0:
        # Run far-field range expert
        cropped_sweep = donut_crop(lidar_sweep)
        far_dets = far_expert(cropped_sweep)
    else:
        # Forecast prev. detections
        far_dets = forecast(dets[time - 1])

    dets[time] = {near_dets, far_dets}
\end{pythonic}

\end{algorithm}

\section{Experiments}
In this section, we demonstrate how detection range affects the accuracy-latency trade off for 3D detectors.
Next, we evaluate a number of popular 3D BEV-based detectors, including PointPillars \cite{lang2019pointpillars}, CBGS\cite{zhu2019class}, CenterPoint \cite{yin2021center}, and TransFusion \cite{bai2022transfusion} on Argoverse 2.0~\cite{wilson2021argoverse}, a long-range detection dataset. 


\subsection{Dataset and Metrics}
\label{sec:setup}

We conduct our experiments on Argoverse 2.0 \cite{wilson2021argoverse}, an autonomous driving dataset with data collected in six US cities. It labels 26 semantic classes for the 3D detection task. Notably, Argoverse 2.0 produces long-range LiDAR point clouds and object annotations (up to $\pm150$m). In comparison, KITTI~\cite{Geiger2012CVPR} only annotates up to $+70$m (with a front facing LiDAR), nuScenes~\cite{caesar2020nuscenes} annotates up to $\pm50$m, and Waymo~\cite{sun2020scalability} annotates up to $\pm75$m. Following standard training protocols used in the nuScenes setup, we adopt 5-frame aggregation for LiDAR densification. We assume that we are provided with ego-vehicle pose for prior frames to align all LiDAR sweeps to the current ego-vehicle pose. Since LiDAR returns are sparse, this densification step is essential for long-range detection.


We evaluate our model using the composite detection score (CDS), a summary metric defined as the product of average precision, computed as an average of four different true-positive thresholds (0.5, 1.0, 2.0, and 4.0 meters) and the sum of the complement of the normalized true positive errors (average translation error (ATE), average scale error (ASE), and average orientation error (AOE)). We refer readers to \cite{Argoverse} for a detailed description of this metric.

\begin{table*}[t]
\small
\centering
\adjustbox{width=\linewidth}{
\setlength{\tabcolsep}{5pt}
\begin{tabular}{@{}cllccccccc@{}}
\toprule
ID & Model & \multicolumn{1}{c}{Method} & Point Proc.      & Backbone  & Neck & Head & Post Proc. \\ \midrule
1  & PointPillars & 50/4 → 50    &   10.5 $\pm$ 3.0         &   3.5 $\pm$ 0.2         &   1.9 $\pm$ 0.1       &   1.2 $\pm$ 0.1        &  58.2 $\pm$ 1.6             \\ 
2  & CBGS &  50/12.5 → 50            &    43.6 $\pm$ 4.0       &   4.7 $\pm$ 0.3         &   2.5 $\pm$ 0.1       &   1.2  $\pm$ 0.2        &   55.9 $\pm$ 3.5             \\
3  & CenterPoint & 50/12.5 → 50     &    45.8 $\pm$  5.5      &   2.7 $\pm$ 0.3       &   0.8 $\pm$ 0.03       &  42.8 $\pm$ 0.6         &   440.9 $\pm$ 48.1           \\
4  & TransFusion-L & 50/12.5 → 50     &   264.9 $\pm$ 45.8         &     4.5 $\pm$ 0.2       &    1.3 $\pm$ 0.3      &   9.8 $\pm$ 3.4        &    1.5 $\pm$ 0.5             \\ \midrule
1  & PointPillars & 100/4 → 100    &   25.6 $\pm$ 7.3         &   10.6 $\pm$ 0.1         &   13.1 $\pm$ 0.1       &   4.1 $\pm$ 0.1        &  62.0 $\pm$ 1.3             \\ 
2  & CBGS &  100/6.25 → 100            &    40.0 $\pm$ 3.2       &   4.7 $\pm$ 0.1         &   2.5 $\pm$ 0.1       &   1.2  $\pm$ 0.1        &   58.6 $\pm$ 1.9             \\
3  & CenterPoint & 100/6.25 → 100     &    42.3 $\pm$  6.1      &   4.8 $\pm$ 0.2       &   0.8 $\pm$ 0.1       &  42.7 $\pm$ 0.6         &   448.1 $\pm$ 54.8           \\
4  & TransFusion-L & 100/6.25 → 100     &    257.7 $\pm$ 34.9                   &      4.5 $\pm$ 0.4                 &    1.3 $\pm$  0.1          &    9.3 $\pm$  2.6       &     1.5 $\pm$  0.2         \\ \midrule
1  & PointPillars & 150/2 → 150    &       4.0  $\pm$ 1.2             &    6.6 $\pm$ 0.3                   &   8.1 $\pm$ 0.2           &   2.7 $\pm$ 0.3        &    60.0 $\pm$ 9.1          \\
2  & CBGS &  150/3.125 → 150           &  35.5 $\pm$ 1.9                    &     3.0 $\pm$ 0.1                  &    1.7 $\pm$ 0.1          &   1.1 $\pm$ 0.1        &   58.8 $\pm$ 1.5           \\
3  & CenterPoint & 150/3.125 → 150     &     33.5 $\pm$ 4.4                 &     3.3 $\pm$ 0.2                  &    0.6 $\pm$ 0.1          &    26.1 $\pm$ 1.0       &   291.1 $\pm$ 66.9           \\    
4  & TransFusion-L & 150/3.125 → 150     &   240.9  $\pm$ 31.6                   &     3.3 $\pm$  0.7                  &     0.8 $\pm$  0.1         &    9.0 $\pm$ 2.1       &      1.5 $\pm$  0.5        \\ \bottomrule

\end{tabular}
} 
\caption{\textbf{Impact of Range on Timing}. We find that increasing range and proportionally decreasing voxel resolution keeps run time (in milliseconds) approximately constant. Within a fixed compute budget, tuning range and voxel resolution are the two key ``knobs'' to trade off latency and accuracy. Further, we find that the point-processing takes a majority of the run time (excluding post-processing). Lastly, we note that CenterPoint's head is more than four times slower than the transformer head in TransFusion and ten times slower than the anchor head in PointPillars. Empirically, we find that this slowdown is due to inefficient bounding box decoding from the CenterPoint regression heads (which can be significanlty optimized). 
}
\label{tab:runtime-breakdown}
\end{table*}
\subsection{Implementation Details}
\label{sec:timing}


 BEV-based detectors often follow the same general architecture. First, sparse LiDAR points are voxelized to form a dense feature map. This dense map is then processed by the SECOND \cite{yan2018second} backbone and FPN neck. Lastly, PointPillars and CBGS process this BEV feature using a minimal SSD-like detection head. CenterPoint uses a center-based detection head which predicts object center's using a heatmap and regresses all other attributes, and TransFusion uses a DETR-like transformer decoder, which directly predicts amodal bounding boxes. All four models predict object semantics and regression bounding box location, size, orientation, and instantaneous velocity. We keep the model architecture fixed for our study, only tuning the range and voxel size. 
We use the open source implementations of these four detectors from mmdetection3d~\cite{mmdet3d2020, bai2022transfusion, peri2022lt3d}. We adopt a basic set of data augmentations, including global 3D tranformations, flip in BEV, and point shuffling during training. We train our model with 8 RTX 3090 GPUs and a batch size of 1 per GPU. The training noise (from random seed and system scheduling) is $<$ 1\% of the accuracy (standard deviation normalized by the mean). We also report the mean timing of three runs for key experiments. For consistent measurement of model runtime, we evaluate with a batch size 1 on a Tesla V100 GPU ~\cite{tan2020efficientdet,Li2020StreamingP,Thavamani2021FOVEA}.

\textbf{Timing of Individual Components:} We provided a detailed component-wise runtime analysis for the models in Table~\ref{tab:runtime-breakdown}. 
CBGS is architecturally identical to PointPillars, but uses a VoxelNet encoder instead of a PillarEncoder. CenterPoint is architecturally identical to CBGS, but uses a center regression head instead of an anchor-based detector head. TransFusion-L (the LiDAR-only variant of TransFusion) is architecturally idential to CenterPoint, but uses a DETR-like transformer decoder as a detector head. Although CBGS, CenterPoint and TransFusion use the same VoxelNet encoder, we find that TransFusion's point processing is nearly 5x slower, likely because it uses spconv1.0 rather than spconv2.0 \cite{spconv2022}. Since the SECOND backbone and neck are identical between the four models, timing numbers are consistent. Further, we note that the anchor-based detector head used in PointPillars and CBGS is the fastest, followed by TransFusion's head. CenterPoint's head is the heaviest, notably taking 40x longer than the anchor-based head. TransFusion's post-processing time is significantly faster than other models because the transformer head decoding stage does not perform non-maximal suppression (NMS). 

In general, post-processing takes a considerable fraction of the runtime for all models. This is a result of using research-level code and can be further optimized, but it is beyond the scope of this work. 
We find that this post-processing time is relatively constant within model-families. For subsequent timing results we omit the post-processing time since this is a dominanting factor which makes analysis more difficult. In practice, this can be sped up using GPU implementations of max-pooled NMS instead of the standard (greedy association) NMS \cite{cai2019maxpoolnms}.


\begin{table}[b]
\small
\centering
\adjustbox{width=\linewidth}{
\setlength{\tabcolsep}{5pt}
\begin{tabular}{@{}cllccccccc@{}}
\toprule
ID & \multicolumn{1}{c}{Method} & 0-50m.      & 50-100m  & 100-150m  \\ \midrule
1  &  50/8 → 50    &   31.5        &          &                          \\ \midrule
2  &  100/4 → 100    &   26.9        &    13.1      &                            \\ 
3  &  + Range Masking    &                 &    9.6      &                              \\ 
4  &  + Up-sampling    &           &   7.7       &                             \\ \midrule
5  &  150/2 → 150   &  16.9      &    9.4      &    5.3                         \\ 
6  &  + Range Masking    &           &          &    3.6                        \\ 
7  &  + Up-sampling   &           &          &   2.1                         \\ \bottomrule
\end{tabular}
} 
\caption{{\bf Range Specialization}. We evaluate range-masking and object up-sampling using PointPillars, and find that both negatively impact the performance of the range-expert, suggesting that the best strategy for training range experts is to generalize to other ranges outside of the region of interest. This conclusion holds for both the $100/4 \rightarrow 100$ and $150 / 2 \rightarrow 150$ range experts. 
}
\label{tab:specialize}
\end{table}

\begin{figure*}[t]
    \centering
    \includegraphics[width=0.44\linewidth]{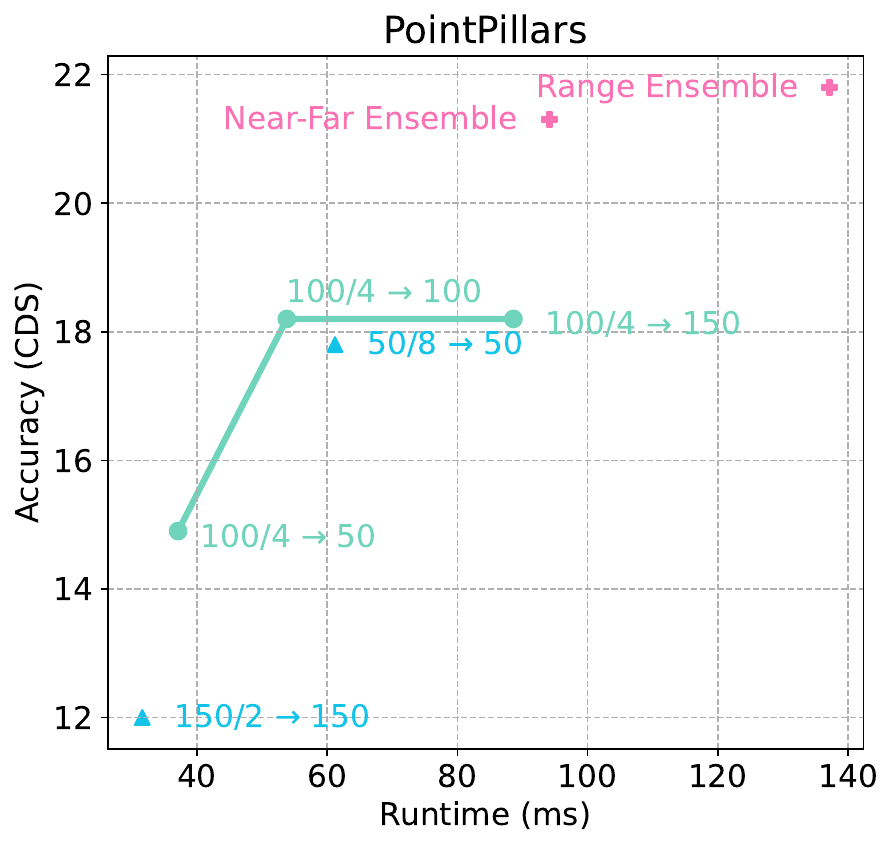}
    \includegraphics[width=0.44\linewidth]{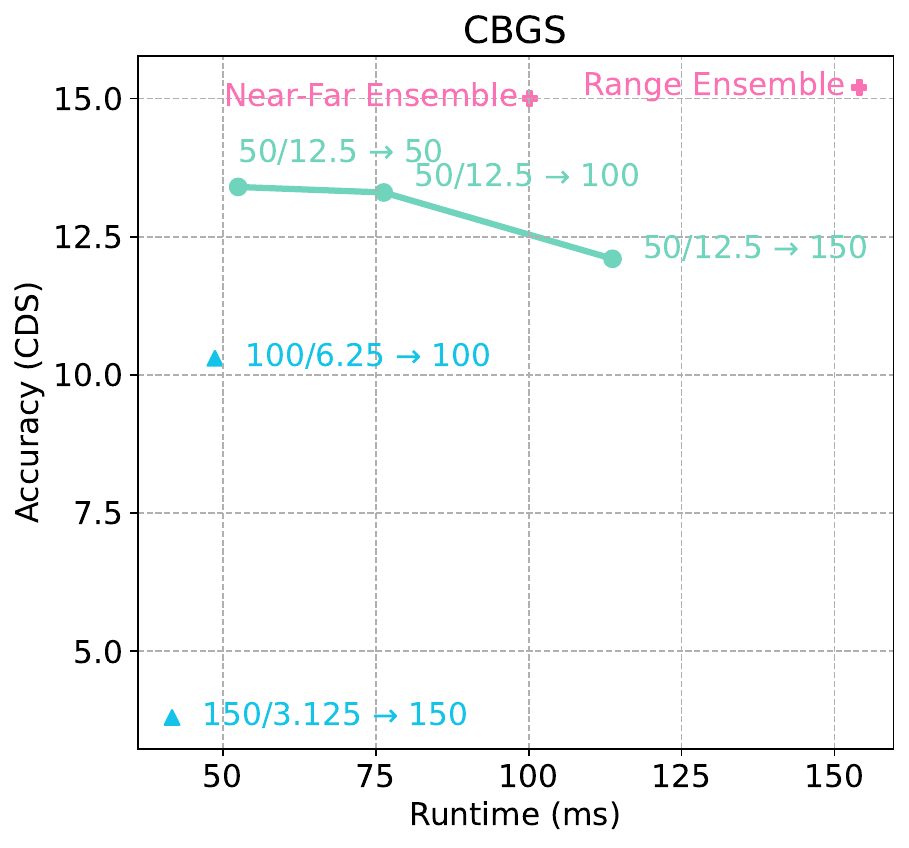} \\
    \includegraphics[width=0.44\linewidth]{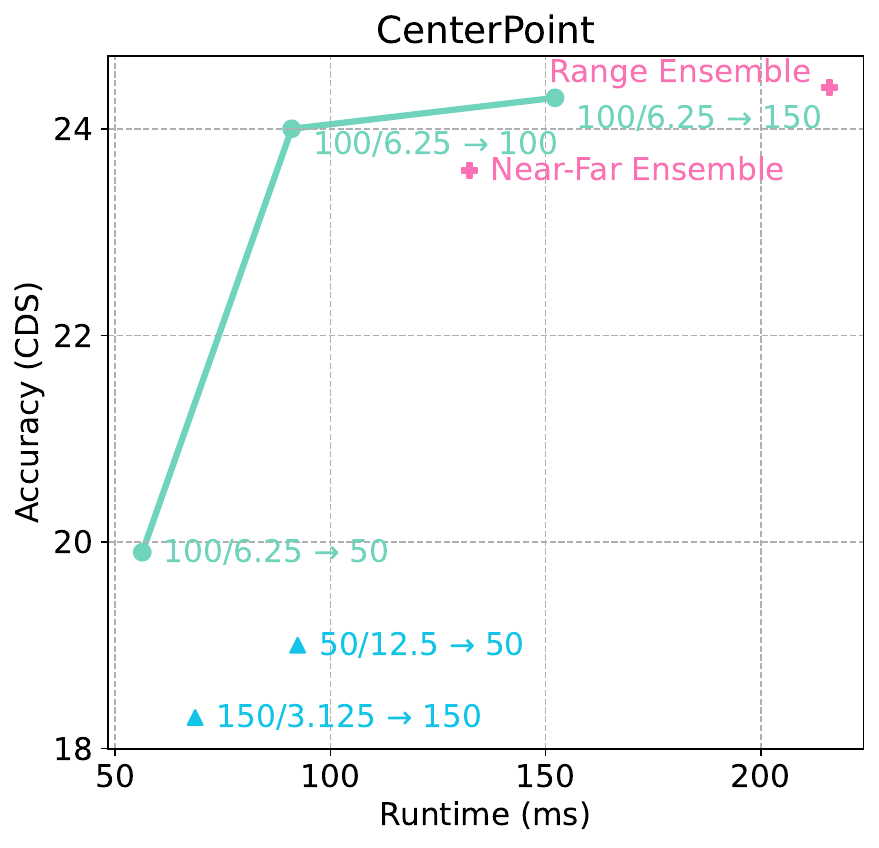}
    \includegraphics[width=0.44\linewidth]{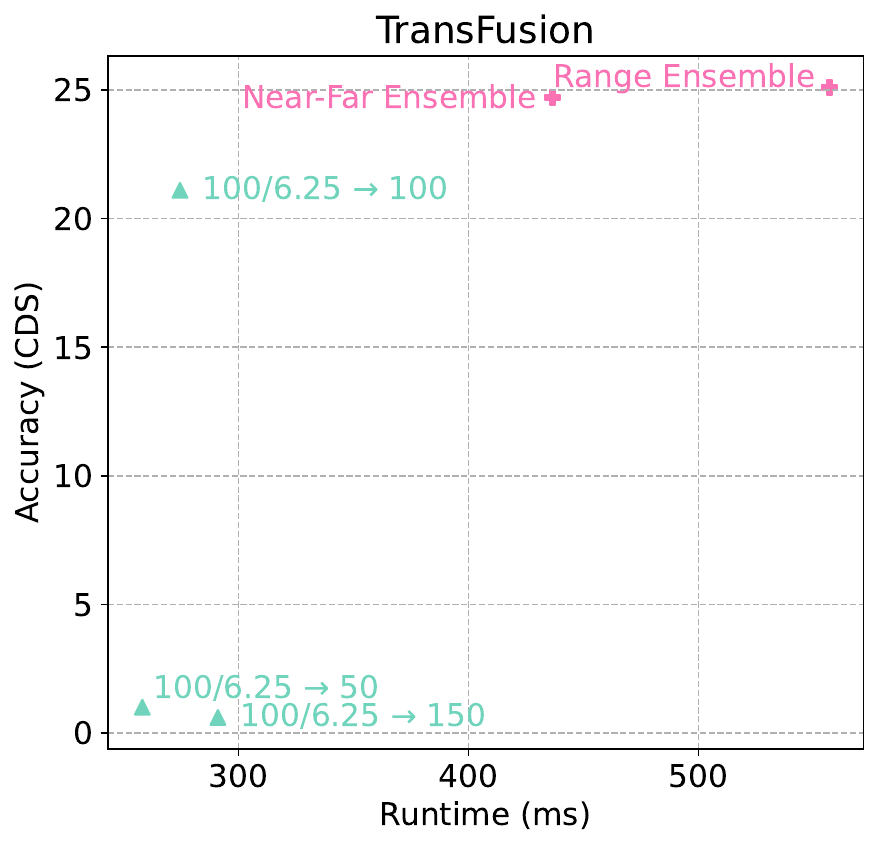}
    \caption{
    Without re-training, we can re-run a 100m trained PointPillars model (top left, $100/4$ green dots) at different ranges (top left, green curve). Interestingly, the 100m PointPillars range expert outperforms both the 50m and 150m range experts (top left, blue triangles), suggesting that we should ``give up'' on far-field detections. However, to avoid giving up on long-range detection, we ensemble our range experts to achieve considerably higher performance than any single PointPillars model, but is almost 3x as slow as the 100m range expert (top left, pink star). In contrast, the near-far ensemble achieves a ``sweet spot'' between the speed of the single range expert and the performance of the naive range ensemble. We see a similar trend with the CenterPoint model family (bottom left). Again, the 100m CenterPoint range expert beats the 50m and 150m range experts. However, the CenterPoint 100m range ensemble evaluated at 150m almost matches the performance of the range ensemble, serving as the best single model. In such cases, we find that the range-ensemble does not provide significant benefit. Unlike the PointPillars and CenterPoint model families, we see that the best-performing CBGS range expert is the 50m model (top right). However, we note that the performance is considerably lower than either PointPillars or CenterPoint. Further, running the 50m CBGS range expert at 100m and 150m leads to degraded performance. This suggests that CBGS does not generalize well to far-field detection. Lastly, we consider TransFusion, a recent state-of-the-art transformer-based detector (bottom right). Unlike the prior three models, we find that we cannot easily run TransFusion in a ``fully-convolutional'' mode. Specifically, although we note strong performance for the 100m range expert when run at 100m, we notice that performance drops to nearly 0 CDS when run on either the 50m or 150m ranges. We posit that TransFusion's use of relative positional encoding rather than metric encoding leads to catastrophically poor across-range generalization. 
    }
    \label{fig:landscape}
\end{figure*}

\subsection{Empirical Analysis}
We evaluate range experts for a variety of detector architectures. Unsurprisingly, we find that the range ensemble consistently outperforms range experts, but is nearly 3x slower. The near-far ensemble provides a ``sweet spot'' that is 33\% faster than the range ensemble, while also often performing better than the best range-expert. Interestingly, we find that some architectures are better at generalizing across-range. 

\textbf{Tuning Range Experts.} We evaluate a simple change to our training protocol which modifies the data distribution to specialize to a particular range interval. As shown in Tab \ref{tab:specialize}, forcing models to specialize to particular ranges by masking out ``donut-holes'' in the point cloud and upsampling far-field objects does not result in better performance compared to the standard approach of training on the full range of LiDAR points. We find that train-time range masking hurts performance, likely because the model is trained with less data overall. For example, in the 50-100m range, the 100m range expert achieves 13.1 CDS whereas the model trained with a ``donut'' shaped LiDAR sweep only attains 9.6 CDS. Similarly, in the 100-150m range, we see that the 150m range expert achieves 5.3 CDS whereas the model trained with a ``donut'' shaped LiDAR sweep gets 3.6 CDS. Similarly, We find that upsampling classes within a specific range interval does worse than the baseline, likely because the distribution of objects seen during train time is significantly different than that seen at test-time. Based on this investigation, range experts should simply train on the full range without specializing for specific range intervals. However, we find that masking out ``donut holes'' in the point cloud during inferences does not affect detection performance, and provides a modest speedup due to sparse computation.

\textbf{Generalization Across Range.} As shown in Figure \ref{fig:landscape}, some models generalize well across ranges (e.g. CenterPoint (bottom right) green curve increases when evaluated on a range beyond the training range), some generalize poorly (e.g. CBGS (top left) green curve decreases when evaluated beyond the training range.), and some don't generalize at all (e.g. TransFusion achieves nearly 0\% CDS when evaluated on a range that is different than the training range). We examine these trends through the lens of model architectures and training losses. 

First, we note that PointPillars has {\em some} generalization capability across ranges. Notably, when we evaluate the 100m range expert at 150m, the performance does not change, indicating that the model likely predicts all far-field detections with a lower confidence than near-field detections.  We argue that knowing what you don't know is a form of generalization. 

Second, as described in Section \ref{sec:timing}, CBGS is architecturally identical to PointPillars, but uses a VoxelNet encoder instead of a PointNet encoder. Unlike PointPillars, CBGS predicted far-field detections with higher confidence than some near-field detections, resulting in lower performance when evaluating long-range detections. We posit that PointPillars' PointNet encoder captures local features that generalized better than the global features encoded by VoxelNet. 

Next, we consider the surprising  across-range generalization capabilities of CenterPoint. CenterPoint is architecturally identical to CBGS, but uses a center regression head instead of an anchor-based detector head. Although these are architecturally similar, we posit that the difference in training loss significantly impacts generalization. CBGS attempts to maximize the IOU of its predictions with the ground truth. In contrast, CenterPoint learns to regress a heatmap of Gaussian targets. We posit that these ``soft targets'' act as a form of data augmentation, which make it easier to train the model. 

 Lastly, we consider TransFusion, which doesn't generalize across-ranges at all.  Specifically, although we observe strong performance for the 100m range expert when evaluating at 100m, we notice that performance drops to nearly 0 CDS when run on either the 50m or 150m ranges. TransFusion-L (the LiDAR-only variant of TransFusion) is architecturally identical to CenterPoint, but uses a DETR-like transformer decoder as a detector head. We posit that using metric positions for positional encoding rather than relative positions may yield better across-range generalisation. 


\section{Conclusion}

We provide analysis on the effect of detection range for 3D object detectors, showing that range is an important ``knob'' to trade off accuracy and latency. We use our analysis to build a simple ensemble of range experts that exploits a fundamental property of LiDAR; namely that sensor returns become sparse at range, allowing for coarser voxel binning. While highly performant, an ensemble of range experts can be slow, and is unsuitable for real-time applications like autonomous navigation. To address this limitation, we propose near-far ensembles, which run near-field detectors at higher frequency (for immediate collision avoidance and far-field detectors at lower frequency (for long-horizon planning). We also explore the generalization of BEV-based 3D detectors across range  and find that certain combinations of voxel encoders and detector heads lead to better across-range generalization.   


\newpage 

{\small
\bibliographystyle{ieee_fullname}
\bibliography{references}

\begin{thebibliography}{10}\itemsep=-1pt

\bibitem{bai2022transfusion}
Xuyang Bai, Zeyu Hu, Xinge Zhu, Qingqiu Huang, Yilun Chen, Hongbo Fu, and
  Chiew-Lan Tai.
\newblock Transfusion: Robust lidar-camera fusion for 3d object detection with
  transformers.
\newblock In {\em Proceedings of the IEEE/CVF Conference on Computer Vision and
  Pattern Recognition}, pages 1090--1099, 2022.

\bibitem{caesar2020nuscenes}
Holger Caesar, Varun Bankiti, Alex~H Lang, Sourabh Vora, Venice~Erin Liong,
  Qiang Xu, Anush Krishnan, Yu Pan, Giancarlo Baldan, and Oscar Beijbom.
\newblock nuscenes: A multimodal dataset for autonomous driving.
\newblock In {\em Proceedings of the IEEE/CVF conference on computer vision and
  pattern recognition}, pages 11621--11631, 2020.

\bibitem{cai2019maxpoolnms}
Lile Cai, Bin Zhao, Zhe Wang, Jie Lin, Chuan~Sheng Foo, Mohamed~Sabry Aly, and
  Vijay Chandrasekhar.
\newblock Maxpoolnms: getting rid of nms bottlenecks in two-stage object
  detectors.
\newblock In {\em Proceedings of the IEEE/CVF Conference on Computer Vision and
  Pattern Recognition}, pages 9356--9364, 2019.

\bibitem{chai2021point}
Yuning Chai, Pei Sun, Jiquan Ngiam, Weiyue Wang, Benjamin Caine, Vijay
  Vasudevan, Xiao Zhang, and Dragomir Anguelov.
\newblock To the point: Efficient 3d object detection in the range image with
  graph convolution kernels.
\newblock In {\em Proceedings of the IEEE/CVF Conference on Computer Vision and
  Pattern Recognition}, pages 16000--16009, 2021.

\bibitem{Argoverse}
Ming-Fang Chang, John~W Lambert, Patsorn Sangkloy, Jagjeet Singh, Slawomir Bak,
  Andrew Hartnett, De Wang, Peter Carr, Simon Lucey, Deva Ramanan, and James
  Hays.
\newblock Argoverse: {3D} tracking and forecasting with rich maps.
\newblock In {\em CVPR}, 2019.

\bibitem{chen2017multi}
Xiaozhi Chen, Huimin Ma, Ji Wan, Bo Li, and Tian Xia.
\newblock Multi-view 3d object detection network for autonomous driving.
\newblock In {\em Proceedings of the IEEE conference on Computer Vision and
  Pattern Recognition}, pages 1907--1915, 2017.

\bibitem{mmdet3d2020}
MMDetection3D Contributors.
\newblock {MMDetection3D: OpenMMLab} next-generation platform for general {3D}
  object detection.
\newblock \url{https://github.com/open-mmlab/mmdetection3d}, 2020.

\bibitem{spconv2022}
Spconv Contributors.
\newblock Spconv: Spatially sparse convolution library.
\newblock \url{https://github.com/traveller59/spconv}, 2022.

\bibitem{fan2021rangedet}
Lue Fan, Xuan Xiong, Feng Wang, Naiyan Wang, and Zhaoxiang Zhang.
\newblock Rangedet: In defense of range view for lidar-based 3d object
  detection.
\newblock In {\em Proceedings of the IEEE/CVF International Conference on
  Computer Vision}, pages 2918--2927, 2021.

\bibitem{Geiger2012CVPR}
Andreas Geiger, Philip Lenz, and Raquel Urtasun.
\newblock Are we ready for autonomous driving? the kitti vision benchmark
  suite.
\newblock In {\em Conference on Computer Vision and Pattern Recognition
  (CVPR)}, 2012.

\bibitem{hu2022point}
Jordan~SK Hu, Tianshu Kuai, and Steven~L Waslander.
\newblock Point density-aware voxels for lidar 3d object detection.
\newblock In {\em Proceedings of the IEEE/CVF Conference on Computer Vision and
  Pattern Recognition}, pages 8469--8478, 2022.

\bibitem{hu2020randla}
Qingyong Hu, Bo Yang, Linhai Xie, Stefano Rosa, Yulan Guo, Zhihua Wang, Niki
  Trigoni, and Andrew Markham.
\newblock Randla-net: Efficient semantic segmentation of large-scale point
  clouds.
\newblock In {\em Proceedings of the IEEE/CVF Conference on Computer Vision and
  Pattern Recognition}, pages 11108--11117, 2020.

\bibitem{Huang2020AnLA}
Rui Huang, Wanyue Zhang, Abhijit Kundu, Caroline Pantofaru, David~A. Ross,
  Thomas~A. Funkhouser, and Alireza Fathi.
\newblock An lstm approach to temporal 3d object detection in lidar point
  clouds.
\newblock In {\em ECCV}, 2020.

\bibitem{jiang2022polarformer}
Yanqin Jiang, Li Zhang, Zhenwei Miao, Xiatian Zhu, Jin Gao, Weiming Hu, and
  Yu-Gang Jiang.
\newblock Polarformer: Multi-camera 3d object detection with polar
  transformers.
\newblock {\em arXiv preprint arXiv:2206.15398}, 2022.

\bibitem{kuang2020voxel}
Hongwu Kuang, Bei Wang, Jianping An, Ming Zhang, and Zehan Zhang.
\newblock Voxel-fpn: Multi-scale voxel feature aggregation for 3d object
  detection from lidar point clouds.
\newblock {\em Sensors}, 20(3):704, 2020.

\bibitem{lang2019pointpillars}
Alex~H Lang, Sourabh Vora, Holger Caesar, Lubing Zhou, Jiong Yang, and Oscar
  Beijbom.
\newblock Pointpillars: Fast encoders for object detection from point clouds.
\newblock In {\em Proceedings of the IEEE/CVF Conference on Computer Vision and
  Pattern Recognition}, pages 12697--12705, 2019.

\bibitem{Li2020StreamingP}
Mengtian Li, Yuxiong Wang, and Deva Ramanan.
\newblock Towards streaming perception.
\newblock In {\em ECCV}, 2020.

\bibitem{Liu2022BEVFusionMM}
Zhijian Liu, Haotian Tang, Alexander Amini, Xinyu Yang, Huizi Mao, Daniela Rus,
  and Song Han.
\newblock Bevfusion: Multi-task multi-sensor fusion with unified bird's-eye
  view representation.
\newblock {\em ArXiv}, abs/2205.13542, 2022.

\bibitem{luo2018fast}
Wenjie Luo, Bin Yang, and Raquel Urtasun.
\newblock Fast and furious: Real time end-to-end 3d detection, tracking and
  motion forecasting with a single convolutional net.
\newblock In {\em Proceedings of the IEEE conference on Computer Vision and
  Pattern Recognition}, pages 3569--3577, 2018.

\bibitem{meyer2019lasernet}
Gregory~P Meyer, Ankit Laddha, Eric Kee, Carlos Vallespi-Gonzalez, and Carl~K
  Wellington.
\newblock Lasernet: An efficient probabilistic 3d object detector for
  autonomous driving.
\newblock In {\em Proceedings of the IEEE/CVF conference on computer vision and
  pattern recognition}, pages 12677--12686, 2019.

\bibitem{peri2022lt3d}
Neehar Peri, Achal Dave, Deva Ramanan, and Shu Kong.
\newblock Towards long tailed 3d detection.
\newblock {\em Conference on Robot Learning}, 2022.

\bibitem{peri2022futuredet}
Neehar Peri, Jonathon Luiten, Mengtian Li, Aljosa Osep, Laura Leal-Taixe, and
  Deva Ramanan.
\newblock Forecasting from lidar via future object detection.
\newblock {\em arXiv:2203.16297}, 2022.

\bibitem{qi2017pointnet}
Charles~R Qi, Hao Su, Kaichun Mo, and Leonidas~J Guibas.
\newblock Pointnet: Deep learning on point sets for 3d classification and
  segmentation.
\newblock In {\em Proceedings of the IEEE conference on computer vision and
  pattern recognition}, pages 652--660, 2017.

\bibitem{qi2017pointnet++}
Charles~Ruizhongtai Qi, Li Yi, Hao Su, and Leonidas~J Guibas.
\newblock Pointnet++: Deep hierarchical feature learning on point sets in a
  metric space.
\newblock {\em Advances in neural information processing systems}, 30, 2017.

\bibitem{rapoport2021s}
Meytal Rapoport-Lavie and Dan Raviv.
\newblock It's all around you: Range-guided cylindrical network for 3d object
  detection.
\newblock In {\em Proceedings of the IEEE/CVF International Conference on
  Computer Vision}, pages 2992--3001, 2021.

\bibitem{shi2019pointrcnn}
Shaoshuai Shi, Xiaogang Wang, and Hongsheng Li.
\newblock Pointrcnn: 3d object proposal generation and detection from point
  cloud.
\newblock In {\em Proceedings of the IEEE/CVF conference on computer vision and
  pattern recognition}, pages 770--779, 2019.

\bibitem{sun2020scalability}
Pei Sun, Henrik Kretzschmar, Xerxes Dotiwalla, Aurelien Chouard, Vijaysai
  Patnaik, Paul Tsui, James Guo, Yin Zhou, Yuning Chai, Benjamin Caine, et~al.
\newblock Scalability in perception for autonomous driving: Waymo open dataset.
\newblock In {\em Proceedings of the IEEE/CVF conference on computer vision and
  pattern recognition}, pages 2446--2454, 2020.

\bibitem{sun2021rsn}
Pei Sun, Weiyue Wang, Yuning Chai, Gamaleldin Elsayed, Alex Bewley, Xiao Zhang,
  Cristian Sminchisescu, and Dragomir Anguelov.
\newblock Rsn: Range sparse net for efficient, accurate lidar 3d object
  detection.
\newblock In {\em Proceedings of the IEEE/CVF Conference on Computer Vision and
  Pattern Recognition}, pages 5725--5734, 2021.

\bibitem{tan2020efficientdet}
Mingxing Tan, Ruoming Pang, and Quoc~V Le.
\newblock Efficientdet: Scalable and efficient object detection.
\newblock In {\em Proceedings of the IEEE/CVF conference on computer vision and
  pattern recognition}, pages 10781--10790, 2020.

\bibitem{tang2020searching}
Haotian Tang, Zhijian Liu, Shengyu Zhao, Yujun Lin, Ji Lin, Hanrui Wang, and
  Song Han.
\newblock Searching efficient 3d architectures with sparse point-voxel
  convolution.
\newblock In {\em European conference on computer vision}, pages 685--702.
  Springer, 2020.

\bibitem{Thavamani2021FOVEA}
Chittesh Thavamani, Mengtian Li, Nicolas Cebron, and Deva Ramanan.
\newblock Fovea: Foveated image magnification for autonomous navigation.
\newblock In {\em ICCV}, 2021.

\bibitem{wang2021scaled}
Chien-Yao Wang, Alexey Bochkovskiy, and Hong-Yuan~Mark Liao.
\newblock Scaled-yolov4: Scaling cross stage partial network.
\newblock In {\em Proceedings of the IEEE/cvf conference on computer vision and
  pattern recognition}, pages 13029--13038, 2021.

\bibitem{wang2019dynamic}
Yue Wang, Yongbin Sun, Ziwei Liu, Sanjay~E Sarma, Michael~M Bronstein, and
  Justin~M Solomon.
\newblock Dynamic graph cnn for learning on point clouds.
\newblock {\em Acm Transactions On Graphics (tog)}, 38(5):1--12, 2019.

\bibitem{wilson2021argoverse}
Benjamin Wilson, William Qi, Tanmay Agarwal, John Lambert, Jagjeet Singh,
  Siddhesh Khandelwal, Bowen Pan, Ratnesh Kumar, Andrew Hartnett,
  Jhony~Kaesemodel Pontes, Deva Ramanan, Peter Carr, and James Hays.
\newblock Argoverse 2: Next generation datasets for self-driving perception and
  forecasting.
\newblock In {\em Thirty-fifth Conference on Neural Information Processing
  Systems Datasets and Benchmarks Track (Round 2)}, 2021.

\bibitem{yan2018second}
Yan Yan, Yuxing Mao, and Bo Li.
\newblock Second: Sparsely embedded convolutional detection.
\newblock {\em Sensors}, 18(10):3337, 2018.

\bibitem{yin2021center}
Tianwei Yin, Xingyi Zhou, and Philipp Krahenbuhl.
\newblock Center-based 3d object detection and tracking.
\newblock In {\em CVPR}, pages 11784--11793, 2021.

\bibitem{zhang2020polarnet}
Yang Zhang, Zixiang Zhou, Philip David, Xiangyu Yue, Zerong Xi, Boqing Gong,
  and Hassan Foroosh.
\newblock Polarnet: An improved grid representation for online lidar point
  clouds semantic segmentation.
\newblock In {\em Proceedings of the IEEE/CVF Conference on Computer Vision and
  Pattern Recognition}, pages 9601--9610, 2020.

\bibitem{zhou2018voxelnet}
Yin Zhou and Oncel Tuzel.
\newblock Voxelnet: End-to-end learning for point cloud based 3d object
  detection.
\newblock In {\em Proceedings of the IEEE conference on computer vision and
  pattern recognition}, pages 4490--4499, 2018.

\bibitem{zhu2019class}
Benjin Zhu, Zhengkai Jiang, Xiangxin Zhou, Zeming Li, and Gang Yu.
\newblock Class-balanced grouping and sampling for point cloud 3d object
  detection.
\newblock {\em arXiv preprint arXiv:1908.09492}, 2019.

\bibitem{zhu2021cylindrical}
Xinge Zhu, Hui Zhou, Tai Wang, Fangzhou Hong, Yuexin Ma, Wei Li, Hongsheng Li,
  and Dahua Lin.
\newblock Cylindrical and asymmetrical 3d convolution networks for lidar
  segmentation.
\newblock In {\em Proceedings of the IEEE/CVF conference on computer vision and
  pattern recognition}, pages 9939--9948, 2021.

\end{thebibliography}
}

\end{document}